\documentclass[runningheads]{llncs}
\usepackage{graphicx}
\usepackage{xcolor}

\usepackage{textcomp}
\usepackage{graphicx}
\usepackage{amsmath,amssymb}
\usepackage{xspace}
\newcommand{\etal}{\emph{et al.}\@\xspace}

\usepackage{array}
\newcolumntype{R}[1]{>{\raggedleft\let\newline\\\arraybackslash\hspace{0pt}}m{#1}}
\usepackage[utf8]{inputenc}
\usepackage{mathtools}
\makeatletter
\def\blfootnote{\xdef\@thefnmark{}\@footnotetext}
\makeatother

\makeatletter
\def\thfootnote{\xdef\@thefnmark{}\@footnotetext}
\makeatother
\usepackage{booktabs,tabu,multirow,tabularx,hhline}

\begin{document}
\title{Weakly-Supervised White and Grey Matter Segmentation in 3D Brain Ultrasound}
\titlerunning{Weakly-Supervised WM and GM Segmentation in 3D Brain Ultrasound}
\author{Beatrice Demiray*\inst{,1} \and Julia Rackerseder*\inst{,1}  \and Stevica Bozhinoski\inst{1} \and Nassir Navab\inst{1,2}}
\authorrunning{Demiray*, Rackerseder* et al.} 
\institute{Technische Universität München, Munich, Germany
\and
Johns Hopkins University, Baltimore, USA}
\maketitle              
\begin{abstract}
\thfootnote{* Both authors contributed equally to this work.}
Although the segmentation of brain structures in ultrasound helps initialize image based registration, assist brain shift compensation, and provides interventional decision support, the task of segmenting grey and white matter in cranial ultrasound is very challenging and has not been addressed yet.
We train a multi-scale fully convolutional neural network simultaneously for two classes in order to segment real clinical 3D ultrasound data. Parallel pathways working at different levels of resolution account for high frequency speckle noise and global 3D image features.
To ensure reproducibility, the publicly available RESECT dataset is utilized for training and cross-validation.
Due to the absence of a ground truth, we train with weakly annotated label. We implement label transfer from MRI to US, which is prone to a residual but inevitable registration error. To further improve results, we perform transfer learning using synthetic US data. 
The resulting method leads to excellent Dice scores of 0.7080, 0.8402 and 0.9315 for grey matter, white matter and background. Our proposed methodology sets an unparalleled standard for white and grey matter segmentation in 3D intracranial ultrasound.
\blfootnote{We received funding from the Horizon 2020 program EDEN2020 under grant agreement No 688279 and by the German Research Foundation (DFG) funded SFB 824. We gratefully acknowledge the support of the GPU grant program from NVIDIA.}
\end{abstract}
\section{Introduction}
\label{sec:introduction}
The standard modality for imaging of brain tissue is magnetic resonance imaging (MRI), where segmentation of grey and white matter (GM and WM) can be performed automatically, and used for many clinical purposes, such as detecting brain abnormalities~\cite{kebir2018efficient} and assessing the severity of dementia~\cite{goyal2017automated}.
Although its advantages have long been proven, intra-operative MRI has not been introduced into many operating theaters and processes, due to time requirements and workflow disruptions.
Ultrasound (US) does not have these constraints; however, images are usually harder to interpret, due to poor contrast, artifacts, low anatomical detail and induced deformations. 
Compared to MRI and CT images, medical US suffers from a limited field of view and speckle noise, caused by scattering of the US beam from tissue inhomogeneities. 
This can lead to decreased contrast between anatomically distinct structures. This reduces the ability of human observers to resolve final detail which complicates and sometimes inhibits expert annotations.
It also impedes robust automated computation of segmentation masks in US.
However, intra-operative US (iUS) is increasingly used in a neurological setting, as a result of ease of integration into the clinical routine and small size.
Thus, it is no longer only used for registration tasks, but also for tissue classification, tracking and intra-op monitoring of patient and surgical processes, drug perfusion and decision support systems.
In this setting, segmentation of brain structures can initialize image based registration ~\cite{rackerseder2018initialize} or assist brain shift and tracking error compensation~\cite{nitsch2019neural} and thus provide interventional decision support.
Nonetheless, segmentation of soft tissue in US, especially intracranial, is a demanding task for algorithms, as well as experts. 
This, combined with the small size of most medical datasets, complicates creating training data of sufficient quality even further.
As a solution labels can be transferred from other modalities, that are easier to annotate or can even be annotated automatically, e.g. CT or MRI.

Although challenging, the segmentation of structures in brain US has extensive relevance in clinical practice and research.
Early diagnosis of Parkinson’s Disease is facilitated by midbrain segmentation from transcranial US (TCUS). Hough-CNN~\cite{milletari2017hough}, automatically segments the midbrain in TCUS and deep brain regions in MRI.
Another active field of research on cranial US segmentation is the automatic detection of ventricles in infants for diagnosis of brain anomalies.
US can spare children sedation for MRI to avoid motion artifacts.
A first fully automatic approach to determine the volume of ventricles~\cite{qiu2017automatic} is based on level-set. 
However, it is difficult to integrate into clinical routine, due to a mean processing time of 54 minutes per volume.
Recent developments in deep learning enable the segmentation of ventricles in infant US. A combination of U-Net and SegNet exploited 2D US~\cite{wang2018automatic}, followed by a solely U-Net based implementation achieving segmentation in 3D US~\cite{martin2018automatic}.
For the segmentation of WM and GM in MRI, standard approaches rely on the fuzzy C-means clustering technique~\cite{goyal2017automated}.
Methods tackling this problem with deep learning include SegNet~\cite{brebisson2015deep} and VoxResNet~\cite{chen2018voxresnet}. 
As a pre-processing step for lesion detection, DeepMedic~\cite{kamnitsas2017efficient} segments GM, WM and ventricles from MRI.
Therefore the segmentation of WM and GM in MRI appears to be a well addressed problem. This is not the case for US. 
%
\begin{figure}[t]
    \includegraphics[width=1.0\textwidth,trim= 10 63 8 45, clip]{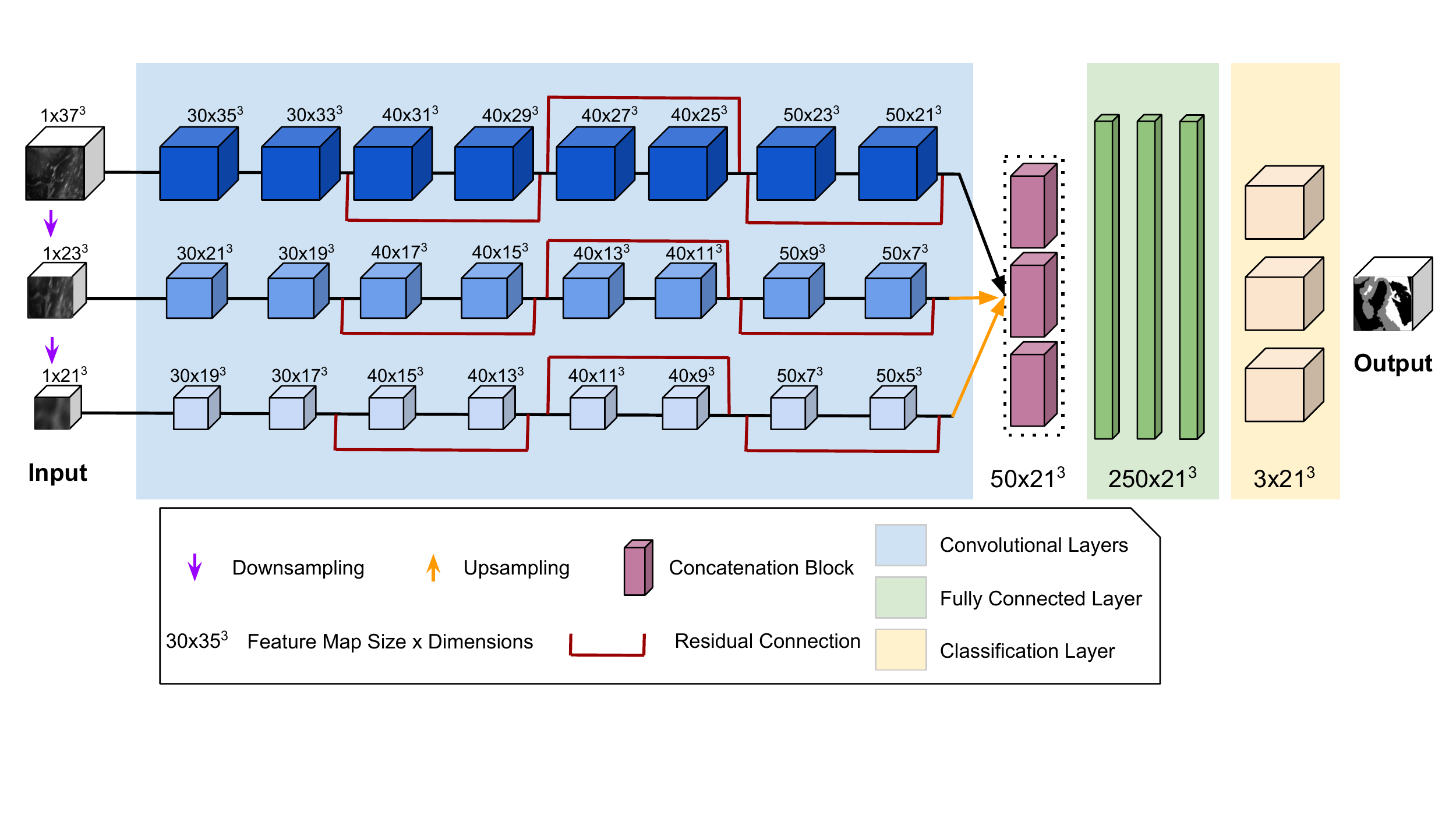}
    \caption{DeepMedicUS with three pathways at different input resolutions.}
    \label{fig:architecture}
\end{figure}
To the best of our knowledge, we present the first work tackling the challenging problem of GM and WM segmentation in 3D US. 
We use and evaluate an extended DeepMedic~\cite{kamnitsas2017efficient} architecture on the publicly available RESECT dataset~\cite{xiao2017retrospective}.
The original network implementation is available online and we provide an elaborate description of all hyperparameters, settings and experimental setups. By using a public dataset, this work allows full reproducibility, easy benchmarking and comparison within the research community. 
To gain insight on how to address the specific appearance of ultrasound data, we perform a study on two different activation functions to evaluate their behavior in Fully Convolutional Networks when applied to US image analysis. We show that pre-training network models on synthetic US data can improve their performance. We present unparalleled results with average dice score of 0.7080, 0.8402 and 0.9315 for GM, WM and background, respectively. Given that ground truth labels are unavailable, we only train with weakly annotated labels which are prone to a residual but inevitable registration error. 
Yet, for some patient cases the model even improves over uncertain transferred labels. 
\section{Methods}
\label{sec:methods}
In this work we propose a CNN designed for the domain of US, inspired by the architecture presented by Kamnitsas~\etal~\cite{kamnitsas2017efficient}. Their segmentation network for MRI outputs label probability maps, regularized by Conditional Random Fields (CRF) to produce highly accurate segmentation labels. 
To account for high frequency speckle present in ultrasound images, as well as more global image features in the 3D US data, we add another parallel pathway, working at an even lower resolution.
In addition, we implemented a cyclic learning rate and empirically adapted the size of the hidden fully convolutional layers to further adapt our network to US.  We train two classes simultaneously with the goal of segmenting multiple anatomical regions in 3D data, acquired in a real clinical setting.
Following, we present the dataset preparation and further elaborate the network architecture.

\textbf{Dataset Preparation}
We utilize T1 weighted MRI and co-registered pre-resection reconstructed 3D US volumes of the public RESECT dataset~\cite{xiao2017retrospective} from 23 patients with low-grade glioma. Patient numbering is kept according to original publication.
Even for clinical experts it is challenging to distinguish GM and WM in US, rendering it impossible to obtain ground truth, which leaves us with the challenge to train our network weakly supervised with the risk of introducing a residual but inevitable registration error.
Thus, for each US volume we generate labelmaps from segmentations in co-registered MRI volumes via label transfer. 
To generate MRI labelmaps, skull stripping and cortical parcellation of MRI volumes are performed automatically with FreeSurfer\footnote{http://surfer.nmr.mgh.harvard.edu/fswiki/}.
The parcellation labelmap is converted to GM~(label $l=1$), WM~($l=2$) and background~(BG, $l=0$) annotations.
MRI and US volumes are co-registered rigidly.
To increase the quality of our propagated labelmaps, we employ an affine registration of MRI and US using the $LC^2$ metric~\cite{fuerst2014automatic} in ImFusion Suite\footnote{Version  1.1.8, ImFusion GmbH, Munich, Germany, https://www.imfusion.com/}.
The annotations for GM, WM and BG are mapped to the US volumes and sampled to a 0.4~mm isotropic resolution. 
This yields a label distribution of 23~\% BG, 30~\% GM and 47~\% WM.

\textbf{Simulated US Sweeps}
To cope with the small dataset size, we evaluate pre-training the network on synthetic US sweeps.
The simulation allows the approximation of different imaging conditions by modifying the imaging parameters and acoustic properties of tissue types. 
Synthetic volumes are generated with a hybrid ray-tracing and convolutional method~\cite{salehi2015patient} based on the MRI labelmaps.
A large amount of sweeps is generated and filtered, resulting in five high quality volumes per patient.

\textbf{Pre-Processing and Data Augmentation}
US volumes and simulated sweeps are resampled to an isotropic voxel size of 0.4~mm.
Standardization is performed by subtracting mean intensity value and dividing by standard deviation, ensuring stable behavior during training.
US volumes are masked to areas that contain image information only.
To tackle the challenges associated with small homogeneous datasets and  encourage convergence to robust models while reducing overfitting, we augment the data per patch during training with a certain probability.
Patches are flipped randomly by one of the main axes 
and rotated by 90° around one arbitrary chosen main axis.
A balanced distribution of foreground and background classes is enforced in a ratio of 1:1 during training.
%
\begin{figure}[ht]
    \includegraphics[width=1.0\textwidth,trim= 157 490 0 75, clip]{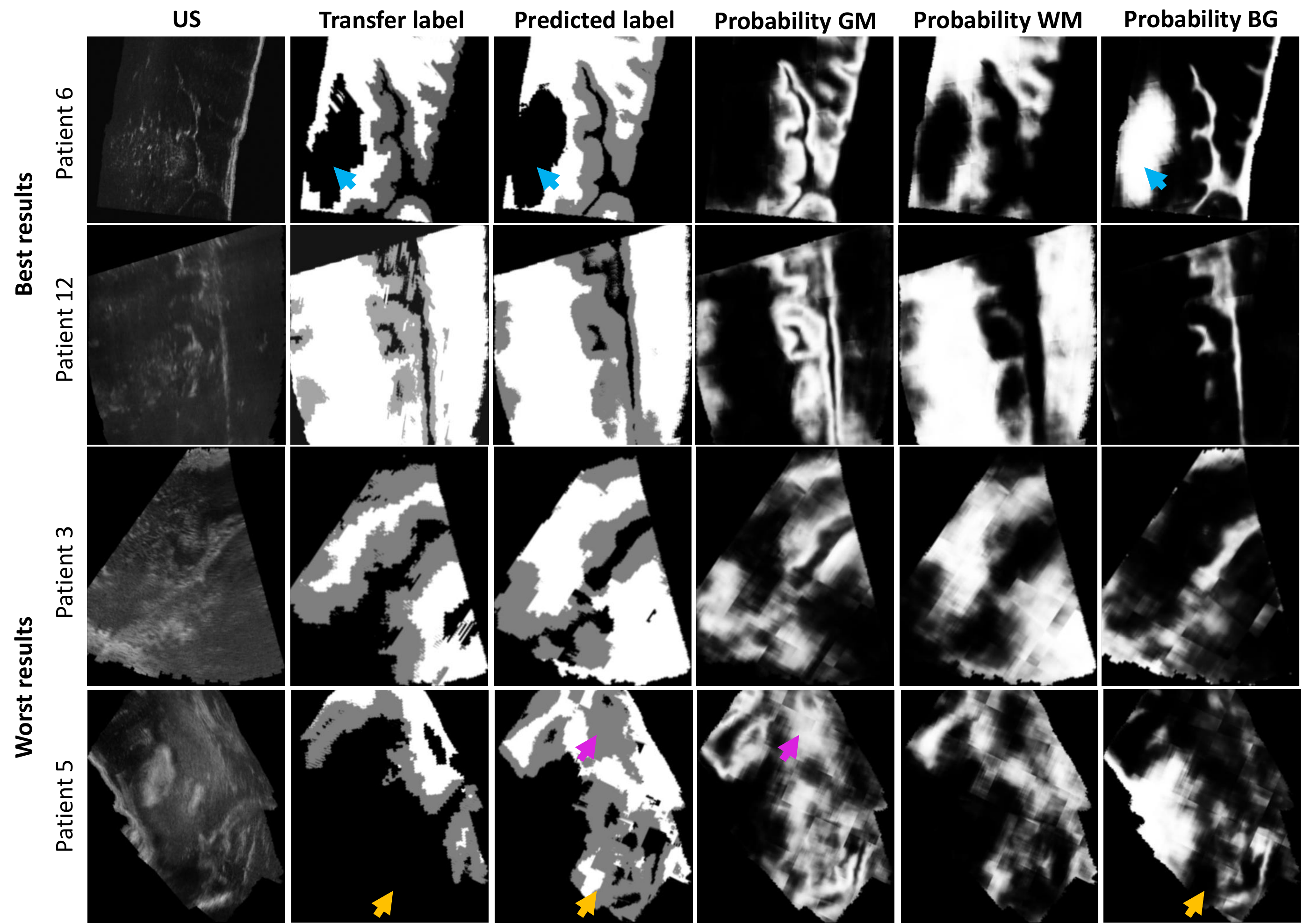}
    \caption{Examples of best and  worst performing patients in the dataset. We show US, label map (WM in white, GM in grey and BG in black), prediction by the network, and probability maps for GM, WM and BG (probability indicated by increasing intensity).
    }
    \label{fig:qualitativeResultsBestVsWorst}
\end{figure}

\textbf{Network Architecture}
Conventional CNNs tend to lose spatial information in their last fully connected layers, which opposes the high importance the information takes in semantic segmentation. 
Fully convolutional networks (FCN) can mitigate this problem by using transposed convolutional layers instead of applying learned up-sampling to low-resolution feature maps. 
We utilize a multi-scale CNN architecture that has achieved promising results for anatomical whole brain~\cite{brebisson2015deep} and lesion segmentation~\cite{moeskops2016automatic} in MR brain images.
As baseline for comparison, the original architecture~\cite{kamnitsas2017efficient} was implemented.
We refer to the proposed network as \textit{DeepMedicUS}. We use a three-pathway approach similar to \cite{kamnitsas2017efficient} with eight convolutional layers per pathway, followed by concatenation blocks and three fully convolutional layers (Fig. \ref{fig:architecture}). 
Batch normalization is applied after each convolutional layer and before each activation layer. 
The first path takes the 3D input patch at the original resolution while the two parallel pathways downsample by factor 3 and 5, respectively. 
This ensures global features are captured while not straining memory. 
The kernel size is set to [3,3,3] except for the two final fully convolutional layers with size [1,1,1].
Interconnecting different network levels preserves high-level image features and speeds up training times.
Thus, residual connections are introduced to layers 4, 6 and 8.
Cross entropy is used as loss function, mini-batch gradient descent with Adam optimizer is used.
To minimize the need for manual refinement of the learning rate, we implement a cyclic learning rate and derived the optimal parameters as described in \cite{smith2017cyclical}. 
This lowers the risk of slow convergence or divergence. 
Tailored to our model, we use a triangular policy with a base learning rate of $1\times10^{-3}$, a maximum bound of $8\times10^{-3}$ and a step size of 1600.
To improve segmentation accuracy, CRF are used as a post-processing step to integrate smoothness terms. 
Thus variable updates can be efficiently executed using Gaussian filtering in feature space to maximize label consistency between similar pixels.

\textbf{Experiments} We analyze the effect of two activation functions: 1)~a rectified linear unit (ReLU), where all negative values are set to zero. 2)~Parametric ReLU (PReLU) which adaptively learns a proper positive slope for negative inputs, preventing negative neurons from dying. For complex US data, we expect this to improve performance at negligible extra computational cost. 
In addition, we analyze the effects of transfer learning using synthetic US data, which is five times the amount of real data. 
For this purpose, we pre-train a model from scratch on the synthetic data.
We then fine-tune and test this model exclusively with real data. To examine the effect of different pre-training dataset sizes, we repeat this experiment at 100~\%, 50~\% and 25~\% of the available synthetic data. 
In order to estimate the performance of our models on unseen data despite having a small dataset, we evaluate all trained models with N-fold cross-validation employing case separation at patient level. 
We randomly separate 23 patient cases into $N=5$ folds containing [5,5,5,4,4] cases, respectively. 
We keep this distribution consistent for all experiments to ensure comparability. For the implementation of all architectures we use the TensorFlow framework. 
All training and testing processes are performed using an NVIDIA TITAN X Pascal GPU (CUDA 10). 
\begin{table}[ht]
\label{tab:resultsDiceAll}
\caption{Comparison of Dice scores for different network architectures and actvation functions. Results are shown per cross validation test fold and per label class.}
    \bgroup
    \setlength{\tabcolsep}{0.2em} 
    \def\arraystretch{1.5} 
    \begin{tabular}{|l|l|c|c|c|c|c|c|c|}
    \hline
    \textbf{Network}                         & \textbf{Setting}                           & $l$     & $n=1$      & $n=2$      & $n=3$ & $n=4$ & $n=5$ & \textbf{Avg$\pm$Std} \\ \hline
    \multirow{ 3}{*}{\textit{Kamnitsas}~\etal}     & \multirow{ 3}{*}{PReLU}     & GM    & 0.5216 & 0.4522 & 0.4376 & 0.5086 & 0.4542 & 0.4749$\pm$0.0376 \\ \cline{3-9}
                                    &                                   & WM    & 0.4899 & 0.4507 & 0.4187 & 0.4874 & 0.4908 & 0.4675$\pm$0.0320 \\ \cline{3-9}
                                    &                                   & BG    & 0.6271 & 0.5832 & 0.4540 & 0.5551 & 0.5986 & 0.5636$\pm$0.0666 \\ \hhline{|=|=|=|=|=|=|=|=|=|}
    \multirow{ 3}{*}{\textit{DeepMedicUS}}    & \multirow{ 3}{*}{PReLU}    & GM    & 0.7025 & 0.7137 & 0.6608 & 0.7105 & 0.7524 & \textbf{0.7080}$\pm$0.0327 \\ \cline{3-9}
                                    &                                   & WM    & 0.8343 & 0.8790 & 0.7885 & 0.8271 & 0.8719 & \textbf{0.8402}$\pm$0.0367 \\ \cline{3-9}
                                    &                                   & BG    & 0.9485 & 0.9557 & 0.8805 & 0.9247 & 0.9480 & \textbf{0.9315}$\pm$0.0308 \\ \hhline{|=|=|=|=|=|=|=|=|=|}
    \multirow{ 3}{*}{\textit{DeepMedicUS}}    & \multirow{ 3}{*}{ReLU}            & GM    & 0.5834 & 0.5234 & 0.5228 & 0.4875 & 0.5897 & 0.5414$\pm$0.0438 \\ \cline{3-9}
                                    &                                   & WM    & 0.7852 & 0.7197 & 0.6495 & 0.6515 & 0.7777 & 0.7167$\pm$0.0656 \\ \cline{3-9}
                                    &                                   & BG    & 0.9115 & 0.8846 & 0.8036 & 0.8569 & 0.8969 & 0.8707$\pm$0.0425 \\ \hline
    \end{tabular}
    \egroup
\end{table}
\section{Results and Discussion }
\label{sec:results}
The quantitative results from our network comparison are shown in Fig.~\ref{tab:resultsDiceAll}. On average, \textit{DeepMedicUS} achieves the highest Dice with PReLU at 0.7080, 0.8402 and 0.9315 for GM, WM and BG, respectively. The additional pathway at lower scale significantly improved the performance of the model tailored for US data.
For GM and WM, specificity (0.8957 and 0.9311) is generally higher than sensitivity (0.8021 and 0.8648), i.e. the number of false negatives is comparatively low which is desirable in clinical applications. While classification of BG pixels appears to be less challenging for the model, for WM and especially GM this poses a more complex task. Over all testing folds, WM predictions show higher accuracy over GM predictions. This could be due to WM structures in general having a more homogeneous appearance and intensity profile, thus being an easier task for the network to identify.
Training the network by Kamnitsas~\etal \cite{kamnitsas2017efficient} took 13.6 hours on one GPU. Training \textit{DeepMedicUS} took 14.7 hours, increasing the computational cost by only $8.1\%$, while achieving more accurate segmentations. PReLU increased computational cost over ReLU by 12.0~\%.  Given the improvement in segmentation accuracy of 30.8~\%, 17.2~\% and 7.0~\% for GM, WM and BG, respectively, this is an acceptable cost.
Segmenting one full patient volume on average took 14 seconds.
We compared the accuracy of pre-trained models at different training dataset size.
A reduction to 25~\% of training data impaired Dice scores by 8~\%, 15~\% and 4~\% for GM, WM and BG, respectively, due to overfitting. Doubling the amount of synthetic data for pre-training, however, plateaued the average accuracy of the model, leading to no improvement in Dice.

Qualitative comparison of the predicted segmentations can be seen in Fig.~\ref{fig:qualitativeResultsBestVsWorst}, which depicts US, labelmaps and predictions for the patients with best and worst results.
For GM the Dice ranges between 0.3964 (patient 5) and 0.8836 (patient 6), for WM 0.3884 (patient 3) and 0.9375 (patient 12) and for BG 0.7011 (patient 5) and 0.9375 (patient 12).
Visual inspection of predictions only leaves few complaints, which is shown in patient 6 and 12.
The network is also able to correctly interpret tumor areas as BG (blue arrows).
These results can be explained partially by tracking inaccuracies that cause reconstruction problems and the use of different US probes, that give worse ultrasound images, see for example patient 3.
For patient 5 the network falsely classifies too much area as GM (purple arrow).
It is able to correctly label an area with incorrect transfer labels as non-BG (yellow arrows), however leading to a low Dice.
Although some volumes seem more challenging for the network, all results are in a clinically acceptable range.

\textit{DeepMedicUS} achieves accurate segmentation with Dice of 0.7080, 0.8402 and 0.9315 for GM, WM and BG, respectively. In comparison, average Dice scores of 91.4~\% for whole tumor segmentation in MRI were reported by \cite{kamnitsas2017efficient}. However, while tumor tissue usually shows good contrast to surrounding healthy tissue in MRI, segmenting anatomical structures inside the tumor was shown to be a more challenging task. Hence, \cite{kamnitsas2017efficient} also report scores of 50.0 and 35.1 for such tasks. This score was further reduced for 50~\% and 20~\% training dataset size. These findings for the influence of data size reduction are coherent with the outcomes presented in our work.
Averaged over all labels, we achieve a comparable Dice of 0.83: Milletari \etal~\cite{milletari2017hough} report an average Dice score of 0.82 for 3D midbrain segmentation in TCUS. For the similar domain of 3D reconstructed transfontanelle ultrasound, \cite{martin2018automatic} report a Dice of 0.816 for the task of cerebral ventricle segmentation.
\section{Conclusion}
\label{sec:Conclusion}
In this work we demonstrated that automatic and robust segmentation of complex anatomical structures in 3D US can be feasible in clinical settings. We validated this on intra-operative cranial US. We used a multi-pathway FCN that is specifically tailored towards the image domain at hand to address the difficult task of segmenting WM and GM in US. We were able to mitigate the problem of losing spatial information and preserve high-level image features. Despite US segmentation generally being a more complex task than MRI segmentation to automate, we achieved good Dice scores. Finally, we could substantiate that pre-training neural networks with synthetic data in the presence of small medical training data can improve the model robustness and accuracy.

%
%
%
\end{document}